\documentclass{article}

\usepackage{PRIMEarxiv}

\usepackage[utf8]{inputenc} 
\usepackage[T1]{fontenc}    
\usepackage{amsmath}
\usepackage{amsfonts}
\usepackage{algorithm}
\usepackage{algorithmic}
\usepackage{graphicx}
\usepackage{xcolor}
\usepackage{subcaption}
\usepackage{comment}
\usepackage{amsfonts}
\usepackage{amssymb}
\usepackage{lineno}
\usepackage{hyperref}
\usepackage{float}
\usepackage{pdflscape}   
\usepackage{multirow}
\usepackage{colortbl}
\usepackage{adjustbox}
\usepackage{longtable}
\usepackage{booktabs}
\usepackage{ltablex}
\usepackage{pdflscape}
\usepackage{cite}
\usepackage[numbers]{natbib}

\title{On the Use of 
Evolutionary Optimization for the Dynamic Chance Constrained Open-Pit Mine Scheduling Problem
}

\author{ \href{https://orcid.org/0000-0002-8043-994X}{\includegraphics[scale=0.06]{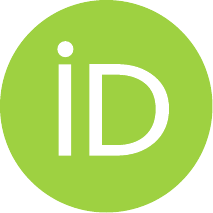}\hspace{1mm}Ishara Hewa Pathiranage} \\
	Machine Learning and Optimisation, \\School of Computer Science and \\Information Technology,\\
  Adelaide University,\\ Adelaide, Australia 
	\And
	\href{https://orcid.org/0000-0002-0036-4782}{\includegraphics[scale=0.06]{orcid.pdf}\hspace{1mm}Aneta Neumann} \\
	Machine Learning and Optimisation, \\
    School of Computer Science and\\ Information Technology,\\
  Adelaide University,\\ 
  Adelaide, Australia \\
}

\begin{document}

\maketitle              
\begin{abstract}
Open-pit mine scheduling is a complex real-world optimization problem that involves uncertain economic values and dynamically changing resource capacities.
Evolutionary algorithms are particularly effective in these scenarios, as they can easily adapt to uncertain and changing environments. However, uncertainty and dynamic changes are often studied in isolation in real-world problems. 
In this paper, we study a dynamic chance-constrained open-pit mine scheduling problem in which block economic values are stochastic and mining and processing capacities vary over time. We adopt a bi-objective evolutionary formulation that simultaneously maximizes expected discounted profit and minimizes its standard deviation. To address dynamic changes, we propose a diversity-based change response mechanism that repairs a subset of infeasible solutions and introduces additional feasible solutions whenever a change is detected. We evaluate the effectiveness of this mechanism across four multi-objective evolutionary algorithms and compare it with a baseline re-evaluation–based change-response strategy.
Experimental results on six mining instances demonstrate that the proposed approach consistently outperforms the baseline methods across different uncertainty levels and change frequencies.
\end{abstract}


\section{Introduction}
The open-pit mine scheduling problem (OPMSP) is a large-scale combinatorial optimization problem that determines the optimal sequence of block extraction from a mining deposit to maximize the net present value (NPV), while satisfying geological and operational constraints. Over the years, OPMSP has been addressed using a variety of optimization approaches, including integer linear programming methods~\cite{Letelier}, heuristics~\cite{JELVEZ20161169}, metaheuristics such as differential evolution~(DE)~\cite{ELSAYED202077}, and evolutionary algorithms~\cite{10612008}, as well as hybrid approaches~\cite{PAITHANKAR2019105507}.

However, real-world mining operations operate under multiple sources of uncertainty and dynamic changes, including stochastic ore grades, commodity price fluctuations, equipment downtime, weather conditions, and time-varying mining and processing capacities. Ignoring these factors can lead to infeasible or suboptimal production schedules, reduced economic returns, and compromised operational safety~\cite{Chatterjee27052020, PAITHANKAR2021101875}.

Several approaches have been proposed to address OPMSP under uncertainty, including exact methods~\cite{Chatterjee27052020, CHATTERJEE2016658} and metaheuristic techniques such as particle swarm optimization~\cite{GILANI2020101738}, differential evolution~\cite{KHAN2018428,10.1145/3449639.3459382}, evolutionary algorithms~\cite{10254112}, and hybrid methods~\cite{PAITHANKAR2019105507}. Nevertheless, to the best of our knowledge, no prior work has simultaneously addressed stochastic profits and dynamically changing resource capacities within the OPMSP framework. Considering both aspects is essential to avoid operational bottlenecks, such as mill overloads or deviations between planned and actual ore grades. Addressing these issues early in the planning process ensures the development of robust, efficient, and operationally feasible mine plans.

Evolutionary algorithms~(EAs) are particularly effective for stochastic and dynamic optimization problems due to their flexibility and robustness~\cite{DBLP:conf/ecai/AssimiHXN020,10.1145/3321707.3321792,10.1145/3638529.3654067,10.1145/3638529.3654081,Roostapour2020}. To handle uncertainty explicitly, chance-constraints~\cite{Charnes} can be used in optimization problems, ensuring that stochastic constraints are satisfied with a predefined confidence level~$\alpha$. In this context, multi-objective evolutionary algorithms~(MOEAs) are particularly effective, as they enable the simultaneous optimization of multiple stochastic components, such as the expected value and variance~\cite{10.1007/s11047-006-9004-x,DBLP:conf/gecco/0001W23}.

In this paper, we investigate the dynamic chance-constrained open-pit mine scheduling problem with stochastic profits and dynamically changing resource capacities.

\subsection{Related Work}

The OPMSP has been extensively studied due to its economic importance and computational
complexity. For instance, a DE-based approach was proposed in~\cite{ELSAYED202077} for
solving the deterministic OPMSP by reducing problem dimensionality via a single-period
formulation. This method incorporates feasibility repair during reproduction and a local
search method, achieving strong performance on instances with up to $112{,}687$
blocks. While early studies typically assume deterministic geological models, more
recent work adopts ensemble-based representations that capture spatial grade
uncertainty through multiple equi-probable realizations generated via geostatistical
simulation.

Geological uncertainty has been explicitly modeled using ensemble block models with
multiple realizations, often combined with chance constraints to limit downside
risk~\cite{Reid0RN21,10254112, PAITHANKAR2021101875}. For example,
\citet{10.1145/3449639.3459382} formulated the stockpile blending problem under grade
uncertainty as a single-objective nonlinear optimization problem with chance
constraints, solved using DE with customized repair operators. Similarly,~\citet{10254112} applied a single-objective EA to the chance-constrained OPMSP,
where uncertainty is incorporated by discounting profits across multiple
realizations. However, both approaches rely on a fixed confidence level, $\alpha$.

Evolutionary algorithms have also shown strong performance on dynamic optimization
problems. Dynamic knapsack, travelling thief problem and submodular optimization problems were studied
in~\cite{Roostapour2020,Roostapour_Neumann_Neumann_Friedrich_2019,DonN024,YanNN24,Neumann2020,neumann2025optimizing}, demonstrating that
MOEAs often outperform single-objective methods. Theoretical analysis of dynamic graph
coloring with incremental changes~\cite{10.1145/3321707.3321792} showed that EA-based
re-optimization is more efficient than restarting from scratch. In real-world
applications, NSGA-II and its variants have been successfully applied to dynamic
hydro-thermal power scheduling~\cite{10.1007/978-3-540-70928-2_60}, effectively tracking
changing Pareto-optimal fronts.

Only a limited number of studies address stochastic and dynamic optimization
simultaneously using EAs. In~\cite{DBLP:conf/ecai/AssimiHXN020}, a bi-objective dynamic
chance-constrained knapsack problem was introduced, where item weights follow
independent uniform distributions and probabilistic constraints are estimated using
Chebyshev’s inequality and the Chernoff bound. The results showed that MOEAs are
particularly effective for dynamic chance-constrained optimization, although separate
runs are required for different confidence levels. \citet{10.1145/3638529.3654067} addressed this limitation using a three-objective formulation, which was independent of
the chosen confidence level. Furthermore,~\citet{10.1145/3638529.3654081} studied both
bi-objective and three-objective formulations with stochastic profits using GSEMO with standard and
sliding-window parent selection, and multi-objective approaches~\cite{PereraNN24} showing clear advantages of these approaches. Recent work on stochastic and dynamic multiple knapsack
problems~\cite{MOEOLSOP} further demonstrates the effectiveness of MOEAs. These studies motivate the application of MOEAs to the dynamic chance-constrained OPMSP considered in this paper.

\subsection{Our Contribution}

To the best of our knowledge, this is the first work to investigate the chance-constrained OPMSP in dynamic environments using MOEAs. We assume that the economic value of each block follows a normal distribution, while mining and processing capacities vary randomly over time. To address this problem, we adopt a bi-objective formulation that simultaneously maximizes the expected net present value and minimizes its standard deviation. This formulation aims to find Pareto-optimal solutions that balance expected profit and risk, subject to chance constraints. By explicitly incorporating risk as a separate objective, the proposed approach eliminates the need to predefine a confidence level during optimization, enabling MOEAs to generate high-quality solutions across a range of confidence levels~$\alpha$ in a single run.

To handle dynamic changes in resource capacities, we introduce a diversity-increasing based change-response mechanism. When a change is detected, a subset of infeasible solutions is repaired using hyper-mutation, and additional randomly generated feasible solutions are inserted to maintain population diversity. This mechanism is integrated into four MOEAs, mutation-only variants of MOEA/D, NSGA-II, SMS-EMOA,
and SPEA2, resulting in MOEA/D-DIV, NSGA-II-DIV, SMS-EMOA-DIV, and SPEA2-DIV, respectively. The proposed approach is evaluated against corresponding re-evaluation–based baseline algorithms, MOEA/D-RE, NSGA-II-RE, SMS-EMOA-RE, and SPEA2-RE, which rely solely on re-evaluation after a change occurs. Experimental results across six mining instances show that the proposed method consistently outperforms the baselines under different uncertainty levels and dynamic change frequencies.

The remainder of the paper is organized as follows. Section~\ref{sec:dcc-opmsp} presents the proposed method for solving the open-pit mine scheduling problem with stochastic profits and dynamic resource constraints. Section~\ref{sec:algorithms_experiments} describes the multi-objective evolutionary algorithms we used and the experimental setup. Next, Section~\ref{sec:experiment_results} presents and analyzes the results. Finally, Section~\ref{sec:conclusion} concludes the paper with key findings.

\section{Open Pit Mine Scheduling Problem with Stochastic Profits and Dynamic Resource Constraints}
\label{sec:dcc-opmsp}

This section defines the dynamic chance-constrained OPMSP, its bi-objective fitness function, and the dynamic change-response mechanism.

\subsection{Problem Definition}

We consider a dynamic chance-constrained open-pit mine scheduling
problem, referred to as the DCC-OPMSP, in which block economic values are stochastic, and resource capacities vary over time.

Let $\mathcal{B}$ denote the set of mineable blocks, and
$\mathcal{T}=\{1,\dots,T\}$ denote the planning periods. A schedule is represented by
binary decision variables
$x=\{x_b^t \mid b \in \mathcal{B},\, t \in \mathcal{T}\}$, where
$x_b^t=1$ if block $b$ is mined in period $t$, and $x_b^t=0$ otherwise.
Let $\mathcal{P}$ denote the set of precedence constraints, where
$(a,b)\in\mathcal{P}$ means that block $a$ must be mined before block $b$.
Let $\mathcal{R}$ be the set of resources. For each resource $r\in\mathcal{R}$,
block $b$ consumes $r_b$ units, and $R_r^t$ denotes the available capacity of
resource $r$ in period $t$. 
Each block $b \in \mathcal{B}$ is associated with a stochastic  profit $p_b$, modeled as a normally distributed random variable,
$p_b \sim \mathcal{N}(\mu_b,\sigma_b^2)$, where $\mu_b$ and $\sigma_b^2$ denote the expected value and variance of block $b$, respectively. Block profits are spatially correlated, reflecting geological dependencies within the orebody. A mining schedule is represented using binary decision variables $x_b^t$, where $x_b^t = 1$ if block $b$ is extracted in period $t$, and $x_b^t = 0$ otherwise. 

For a given solution $x$, the total discounted NPV is a random variable defined as
$
p(x) = \sum_{t \in \mathcal{T}} \frac{1}{(1 + d)^t}
\sum_{b \in \mathcal{B}} p_b \, x_b^t$,
where $d$ denotes the discount rate per annum. Since $p(x)$ is a linear combination of (possibly correlated) normally distributed random variables, it also follows a normal distribution.

The DCC-OPMSP is formulated as follows:
\begin{align}
\textbf{Maximize} \quad & P, \nonumber\\
\textbf{Subject to} \quad 
& \Pr\!\left(p(x) \geq P\right) \geq \alpha, \label{eq:dcc_chance} \\
& \sum_{\tau=1}^{t} x_a^\tau \geq \sum_{\tau=1}^{t} x_b^\tau,
\quad \forall (a,b) \in \mathcal{P}, \; t \in \mathcal{T}, \label{eq:dcc_prec} \\
& \sum_{t \in \mathcal{T}} x_b^t \leq 1,
\quad \forall b \in \mathcal{B}, \label{eq:dcc_once} \\
& \sum_{b \in \mathcal{B}} r_b \, x_b^t \leq R_r^t,
\quad \forall r \in \mathcal{R}, \; t \in \mathcal{T}, \label{eq:dcc_resource} \\
& x_b^t \in \{0,1\},
\quad \forall b \in \mathcal{B}, \; t \in \mathcal{T}. \label{eq:dcc_binary}
\end{align}

The objective is to find a schedule $x$ such that the probability of achieving a discounted NPV of at least $P$ is no less than a confidence level $\alpha$. Constraint~\eqref{eq:dcc_chance} enforces the chance constraint, while Constraints~\eqref{eq:dcc_prec}–\eqref{eq:dcc_resource} ensure precedence feasibility, single extraction of blocks, and compliance with mining and processing capacity limits, respectively.  Constraint~\eqref{eq:dcc_binary} ensures the binary nature of the decision variables. We assume $\alpha \in [1/2,1)$ in this paper.

\begin{figure}[!t]
    \centering
    \includegraphics[width=0.7\textwidth]{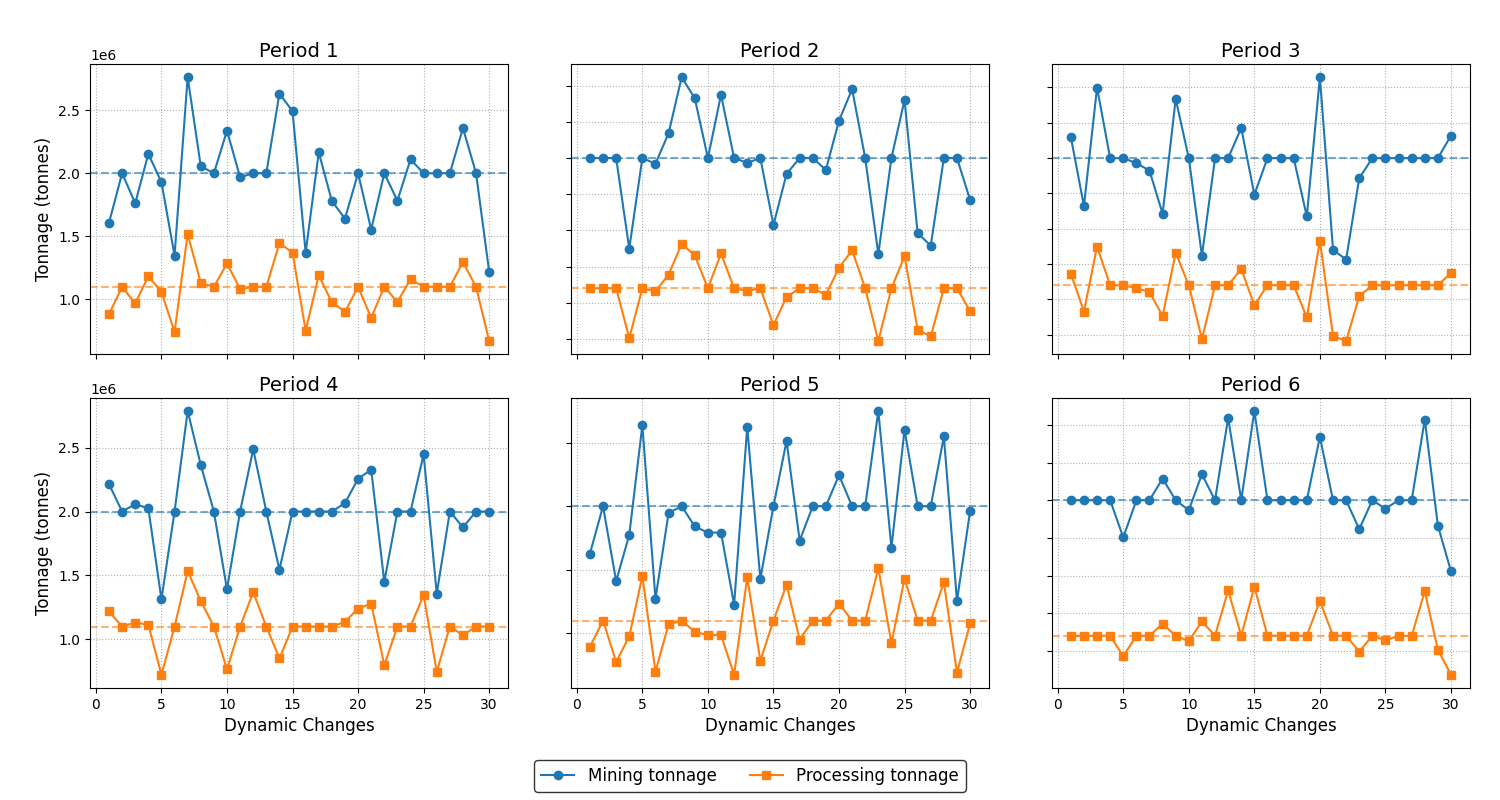}
    \caption{Dynamic changes in mining and processing capacities over six periods
    under 30 dynamic changes. Each subplot represents one period, showing deviations
    from the baseline capacity limits (dashed lines) for the Newman1 instance.}
    \label{fig:dynamic_capacity_changes}
\end{figure}

In addition to stochastic block profits, we incorporate dynamic changes in
resource capacities over time. Every $\tau$ fitness evaluations, we randomly
select a subset of periods and apply capacity variations. For each selected
period, we sample a scaling factor $\gamma \sim \mathcal{U}(1-\eta,1+\eta)$ and
scale the baseline capacities of all resources in that period by $\gamma$. Figure~\ref{fig:dynamic_capacity_changes} illustrates an example of such dynamic capacity variations for the Newman1 instance~\cite{Espinoza2013}, where mining and processing capacities are shown in blue and orange, respectively. Solid lines represent modified capacity limits following dynamic changes, while dashed lines indicate the baseline capacities.

\subsection{Chance-Constrained Profit Modeling}

We now describe how the chance constraint in Constraint~\eqref{eq:dcc_chance} is
modeled and evaluated. We adopt the chance-constrained optimization framework for the knapsack problem
with normally distributed profits proposed in~\cite{DBLP:conf/ppsn/NeumannXN22},
which allows the probabilistic constraint to be reformulated as a deterministic,
risk-adjusted objective.

Accordingly, the objective function takes the following form:
\begin{equation}
f(x) = \mathbb{E}[p(x)] - K_\alpha \cdot \sqrt{\mathrm{Var}[p(x)]},
\label{eq:cc_objective}
\end{equation}
where $K_\alpha = \Phi^{-1}(\alpha)$ denotes the $\alpha$-quantile of the standard
normal distribution.

The expected discounted profit is given by
\[
\mathbb{E}[p(x)] =
\sum_{t \in \mathcal{T}} \frac{1}{(1 + d)^t}
\sum_{b \in \mathcal{B}} \mu_b \, x_b^t .
\]

Let $\mathcal{B}^{\text{ore}} \subseteq \mathcal{B}$ denote the set of ore blocks,
i.e., blocks that generate stochastic processing profit. Since each block is extracted at most once and its uncertainty is block-specific,
the variance of the total discounted profit is defined as
\[
\mathrm{Var}[p(x)] =
\sum_{t \in \mathcal{T}} \frac{1}{(1 + d)^{2t}}
\, \mathrm{Var}_t(x),
\]
where $\mathrm{Var}_t(x)$ denotes the variance of the total profit obtained in
period~$t$. Following~\cite{10254112}, the period-wise variance is computed as
\begin{equation}
\mathrm{Var}_t(x) =
\sum_{b \in \mathcal{B}^{\text{ore}}} \sigma_b^2 \, x_b^t
+
\max\!\left(
0,
\sum_{\substack{b,b' \in \mathcal{B}^{\text{ore}} \\ b \neq b'}}
\mathrm{Cov}(b,b') \, x_b^t x_{b'}^t
\right),
\end{equation}
where $\sigma_b^2$ denotes the variance of the profit of ore block $b$, and
$\mathrm{Cov}(b,b')$ denotes the covariance between the profits of ore blocks
$b$ and $b'$. The variance and covariance terms are estimated using ensemble-based
profit realizations following~\cite{10254112}. To remain consistent with the
modeling assumptions in~\cite{10254112} and to avoid underestimating risk due to
sampling noise, negative covariance values are truncated to zero.

\subsection{Bi-Objective Fitness Function}

We employ a bi-objective formulation for the DCC-OPMSP, defined as
\begin{equation}
g_{2D}(x) = \big(f_1(x), f_2(x)\big),
\end{equation}
where $f_1(x)$ maximizes the expected discounted net present value, while $f_2(x)$ minimizes its standard deviation.

The objectives are defined as follows.
\begin{equation}
f_1(x) =
\begin{cases}
\mathbb{E}[p(x)], & \text{if } v(x) = 0, \\
-\,v(x), & \text{otherwise,}
\end{cases}
\end{equation}
\begin{equation}
f_2(x) =
\begin{cases}
\sqrt{\mathrm{Var}[p(x)]}, & \text{if } v(x) = 0, \\
\mathrm{Var}[p(x)] + M \cdot v(x), & \text{otherwise,}
\end{cases}
\end{equation}
where $\mathbb{E}[p(x)]$ denotes the expected discounted NPV and $\mathrm{Var}[p(x)]$ is the total variance of the discounted NPV
across all periods. The constant $M$ is a large penalty, and
$v(x)$ represents the total penalty due to resource violations.
The total violation is computed as
$v(x) = \sum_{t=1}^{T} v_t(x)$, where the violation in period $t$ is defined as
$v_t(x) = \max_{r \in \mathcal{R}} \max\big(0,\, y_r^t - R_r^t \big)$,
where $y_r^t = \sum_{b \in \mathcal{B}} r_b x_b^t$ denotes the usage of resource $r$
in period $t$ and $R_r^t$ is its corresponding capacity limit.

This bi-objective formulation avoids fixing a confidence level $\alpha$ during
optimization because both objectives, $f_1(x)$ and $f_2(x)$, are independent of
$\alpha$. As a result, the evolutionary algorithm explores a range of
risk-return trade-offs in a single run. After optimization, we can evaluate the
risk-adjusted discounted objective value (Eq.~\eqref{eq:cc_objective}) at
different confidence levels and select appropriate schedules without
re-optimizing.

\subsection{Dynamic Change Response Mechanism with Diversity Increase}

We modify the resource capacities of a randomly selected subset of periods
every $\tau$ iterations. When a change occurs, the population is re-evaluated
under the updated constraints. However, re-evaluation alone is often
insufficient, especially when the population has already converged to locally
optimal or infeasible regions~\cite{10.1145/3524495}.

\begin{algorithm}[!t]
\caption{Dynamic Change Response Mechanism with Diversity Increase}
\label{algo:dynamic_handle}
\begin{algorithmic}[1]
\REQUIRE Population $P$ of size $N$, mutation operator $M$
\STATE $P \gets \{\text{Evaluate}(s)\mid s\in P\}$
\STATE $F \gets \{s\in P \mid v(s)=0\}$, \quad $I \gets P \setminus F$
\IF{$|F| < N$}
    \STATE $S \gets \text{RandomSelect}(I, \min\{0.2N, |I|\})$
    \STATE $F \gets F \cup \{\text{Repair}(s,M)\mid s \in S\}$
    \STATE $k \gets \min\{0.2N,\,N-|F|\}$
    \STATE $F \gets F \cup \{s_1,\dots,s_k\}$ where $s_i \sim \text{RandomFeasible}$
    \WHILE{$|F| < N$}
        \STATE $F \gets F \cup \{\arg\min_{u \in I \setminus F} v(u)\}$
    \ENDWHILE
\ENDIF
\STATE $P \gets F$
\end{algorithmic}
\end{algorithm}
We propose a diversity-increasing change response mechanism for the
DCC-OPMSP~(Algorithm~\ref{algo:dynamic_handle}). After a dynamic change,
the population is re-evaluated and partitioned into a feasible set $F$ and
an infeasible set $I$. If $|F| < N$, feasibility and diversity are restored
in two steps. First, up to $20\%$ of the population is selected from $I$
for repair, bounded by $|I|$, using the hypermutation-based operator~(Algorithm~\ref{algo:repair}). In this process, solutions are accepted based
on feasibility and lexicographic comparison, prioritizing $f_1$ over $f_2$,
or by minimizing constraint violation when infeasible. Second, up to $20\%$
of the population is replaced with randomly generated feasible solutions,
bounded by the remaining capacity $N - |F|$.
\begin{algorithm}[!t]
\caption{Repair Mechanism with Hypermutation}
\label{algo:repair}
\begin{algorithmic}[1]
\REQUIRE Initial solution $s$, mutation operator $M$, mutation probability $P_m$
\STATE $s_{\text{best}} \gets s$, \quad $P_m \gets 2P_m$
\WHILE{stopping criterion not met}
    \STATE $y \gets \text{Evaluate}(M(s_{\text{best}}))$
    \IF{$v(y)=0$ \AND $\big(v(s_{\text{best}})>0 \lor (f_1(y),f_2(y)) \prec_{\text{lex}} (f_1(s_{\text{best}}),f_2(s_{\text{best}}))\big)$}
        \STATE $s_{\text{best}} \gets y$
    \ELSIF{$v(y)>0$ \AND $v(y)<v(s_{\text{best}})$}
        \STATE $s_{\text{best}} \gets y$
    \ENDIF
\ENDWHILE
\RETURN $s_{\text{best}}$
\end{algorithmic}
\end{algorithm}

\section{Multi-Objective Evolutionary Algorithms and Experimental Setting}
\label{sec:algorithms_experiments}
In this section, we present the multi-objective evolutionary algorithms used and the experimental setting.

\subsection{Multi-Objective Evolutionary Algorithms}

We evaluate the proposed DCC-OPMSP using four widely used
multi-objective evolutionary algorithms: the
Multi-Objective Evolutionary Algorithm based on Decomposition (MOEA/D)~\cite{4358754},
the Non-dominated Sorting Genetic Algorithm II (NSGA-II)~\cite{Deb2002AFA},
the S-Metric Selection Evolutionary Multi-Objective Algorithm (SMS-EMOA)~\cite{BEUME20071653},
and the Strength Pareto Evolutionary Algorithm 2 (SPEA2)~\cite{ziztler2002spea2}.
These algorithms have different selection and diversity preservation mechanisms
and are commonly used as baselines in multi-objective optimization under uncertainty
and dynamic environments.

MOEA/D decomposes the multi-objective problem into scalar subproblems using a
set of weight vectors. Each solution is associated with a weight vector,
and neighborhoods are defined based on the distances between weight vectors. We use
the Tchebycheff aggregation function
$g^{te}(x \mid \lambda, z^{*}) =
\max_i \left\{ \lambda_i \lvert f_i(x) - z_i^{*} \rvert \right\}$,
where $\lambda = (\lambda_1,\ldots,\lambda_m)$ is the weight vectors, satisfying $\lambda_i \geq 0$ and
$\sum_{i=1}^m \lambda_i = 1$, $f_i(x)$ denotes the $i$-th objective function, and
$z^{*}$ denotes the current reference (ideal) point.

NSGA-II maintains a population of size $N$ and applies non-dominated sorting and
crowding distance to rank solutions in the population. At each generation, an offspring population of
size $N$ is generated using binary tournament selection and mutation, and the best
$N$ solutions from the combined parent and offspring populations are selected. 

SPEA2 maintains a population and an external archive of non-dominated solutions.
Fitness values are computed using dominance strength and a density estimate based on
the $k$-th nearest neighbor. The archive is updated at each generation to preserve
high-quality and well-distributed solutions. 

SMS-EMOA directly maximizes the hypervolume indicator. At each iteration, one offspring
solution is generated and inserted into the population, and the solution with the
smallest hypervolume contribution is removed.

For dynamic environments, we extend each algorithm with the proposed
diversity-based change response mechanism, resulting in MOEA/D-DIV, NSGA-II-DIV,
SMS-EMOA-DIV, and SPEA2-DIV. As dynamic baselines, we also consider a
re-evaluation-based change-response method, denoted as MOEA/D-RE, NSGA-II-RE,
SMS-EMOA-RE, and SPEA2-RE.

\subsection{Experimental Setting}

We use six standard benchmark instances based on real-world mining projects and simulated data 
from the MineLib library~\cite{Espinoza2013}.
Specifically: Newman1, Zuck Small, Zuck Medium, Marvin are conceptually simulated mines, KD represents a copper deposit in Arizona, USA, and P4HD is a gold and copper deposit located in Nevada, USA. Table~\ref{tab: dataset} summarizes the main
characteristics of these instances.
\begin{table}[!t]
\centering
\caption{Summary of problem instances from MineLib~\cite{Espinoza2013}.}
\label{tab: dataset}
\begin{tabular}{lrrrrrr}
\hline
\textbf{Instance} & \textbf{T} & \textbf{Blocks} & \textbf{Predec.} &
\textbf{Vars.} & \textbf{Res.} & \textbf{Disc.} \\
\hline
Newman1 & 6 & 1\,060 & 3\,922 & 6\,360 & 2 & 0.08 \\
Zuck Small & 20 & 9\,400 & 145\,640 & 188\,000 & 2 & 0.10 \\
KD & 12 & 14\,153 & 219\,778 & 169\,836 & 1 & 0.15 \\
Zuck Medium & 15 & 29\,277 & 1\,271\,207 & 439\,155 & 2 & 0.10 \\
P4HD & 10 & 40\,947 & 738\,609 & 409\,470 & 2 & 0.00 \\
Marvin & 20 & 53\,271 & 650\,631 & 1\,065\,420 & 2 & 0.10 \\
\hline
\end{tabular}
\end{table}

We model the block economic values as normally distributed random variables,
consistent with prior work~\cite{pathiranage2025,10254112}. Geological uncertainty
is represented using $50$ spatially correlated ensemble realizations, indexed by
$e \in \mathcal{E} = \{1,\dots,50\}$. For the Newman1 and Marvin instances, we generate grade ensembles to represent uncertainty. For the remaining instances, where explicit grade data is unavailable, 
we generate ensemble profit
realizations $p_{be}$ for each block $b \in \mathcal{B}$ and ensemble member
$e \in \mathcal{E}$. The standard deviation of each block profit distribution is
set to $20\%$ of its mean value. Chance constraints are evaluated at confidence
levels $\alpha \in \{0.60, 0.90, 0.99\}$.

In dynamic settings, we consider different numbers of dynamic changes during
optimization, $\nu \in \{20, 10, 5, 2\}$, where larger values of $\nu$ correspond to
more frequent changes. Dynamic changes are introduced periodically every $\tau$ fitness evaluations,
where $\tau$ is determined by the total evaluation budget $E_{\max}$ as
$\tau = E_{\max} / \nu$. The magnitude of each dynamic change is controlled by
$\eta = 0.4$ through the scaling factor $\gamma$.

We set the population size to $20$ for
all algorithms. Following~\cite{10.1145/3449639.3459382},  we allocate $10{,}000$ fitness evaluations per run.
We use a greedy–randomized heuristic for initial feasible solution
generation that assigns high-value blocks and their predecessors to earlier periods while respecting precedence and resource constraints, and then improve it using a domain-specific period swap mutation with probability
$P_m = 0.1$ that reassigns blocks across periods, favoring early extraction of ore and delayed removal of waste, without violating feasibility~\cite{pathiranage2025}. 

For MOEA/D-RE and MOEA/D-DIV, we use the neighborhood size
$8$, neighborhood selection probability $0.9$, and replacement of up to $12$
neighboring solutions per iteration. NSGA-II-RE, NSGA-II-DIV, SPEA2-RE, and SPEA2-DIV employ
equal-sized parent and offspring populations, while SMS-EMOA-RE and
SMS-EMOA-DIV generate a single offspring per iteration.

We evaluate the performance of algorithms using the offline error, which measures the
deviation between the best solution found after each change and the deterministic
upper bound $p(x^*)$, obtained by solving the relaxed deterministic OPMSP using the
Gurobi Optimizer~\cite{gurobi}.
For each change $i$, let $x_i$ denote the best solution obtained after the $i$-th change.
If $x_i$ is feasible, the offline error is defined as $e_i = p(x^*) - p(x_i)$.
Otherwise, we penalize infeasibility by setting $e_i = p(x^*) + v(x_i)$, where
$v(x_i)$ is the total resource-violation measure defined in Section~\ref{sec:dcc-opmsp}.
The overall offline error is computed as
$E = \frac{1}{\nu} \sum_{i=1}^{\nu} e_i$.
We use the Kruskal-Wallis test with a $95\%$
confidence level and Bonferroni post-hoc correction~\cite{Corder_Foreman_2014} to evaluate the statistical significance. For the pairwise comparisons, $X^{(+)}$ indicates that the algorithm in the column outperforms algorithm $X$, $X^{(-)}$ indicates the opposite, and $X^{(*)}$ denotes no statistically significant difference.

\section{Experimental Results}
\label{sec:experiment_results}
This section evaluates the proposed change response mechanism on the DCC-OPMSP against re-evaluation baselines. Tables~\ref{tab:offline_error_mean_std} and \ref{tab:offline_error_stats_all} present the mean, standard deviation, and statistical comparisons of offline error across six mining instances under stochastic and dynamic settings. 

Lower mean offline error values correspond to better performance, indicating that the solutions achieve profits closer to the optimal profit $p({x_i}^*)$. All algorithms with the diversity-based change response method (-DIV) consistently show lower mean offline error than their baseline counterparts (-RE), indicating that they adapt more effectively to the changes. This suggests that re-evaluation-based algorithms often struggle to recover feasibility quickly after a change, particularly when dynamic changes are frequent. For all settings, it is shown that when the confidence level of the chance constraint becomes tighter, the offline error increases. Higher confidence levels restrict feasible solutions, requiring more risk-averse extraction sequences. Therefore, the best obtained profit decreases, which tends to increase the offline error. When $\nu=20$, MOEA/D-DIV outperforms other algorithms with a diversity-based mechanism for all the mining instances, while SPEA2-DIV underperforms other -DIV algorithms. However, when $\nu=2$, almost all the -DIV variants perform similarly. In this setting, MOEA/D-RE also performs similarly to MOEA/D-DIV for the Zuck Small instance. Overall, the offline error decreases as the number of dynamic changes $\nu$ decreases,
since less frequent changes allow the algorithms more time to adapt and converge
to feasible solutions. 
\begin{table*}[htbp]
\centering
\footnotesize
\caption{Mean and standard deviation of offline error ($\times 10^{6}$) under different dynamic change frequencies ($\nu$) and confidence levels ($\alpha$). The best mean across all algorithms is highlighted in gray, and pairwise comparisons between (–RE) and (–DIV) variants are highlighted in bold.}
\label{tab:offline_error_mean_std}
\resizebox{\textwidth}{!}{%
\begin{tabular}{lllcccccccccccccccc}
\hline
\multirow{2}{*}{\textbf{Inst.}} & \multirow{2}{*}{\textbf{$\nu$}} & \multirow{2}{*}{\textbf{$\alpha$}} & \multicolumn{2}{c}{\textbf{MOEA/D-RE\textsuperscript{(1)}}} & \multicolumn{2}{c}{\textbf{MOEA/D-DIV\textsuperscript{(2)}}} & \multicolumn{2}{c}{\textbf{NSGA-II-RE\textsuperscript{(3)}}} & \multicolumn{2}{c}{\textbf{NSGA-II-DIV\textsuperscript{(4)}}} & \multicolumn{2}{c}{\textbf{SPEA2-RE\textsuperscript{(5)}}} & \multicolumn{2}{c}{\textbf{SPEA2-DIV\textsuperscript{(6)}}} & \multicolumn{2}{c}{\textbf{SMS-EMOA-RE\textsuperscript{(7)}}} & \multicolumn{2}{c}{\textbf{SMS-EMOA-DIV\textsuperscript{(8)}}} \\
\cline{4-5}\cline{6-7}\cline{8-9}\cline{10-11}\cline{12-13}\cline{14-15}\cline{16-17}\cline{18-19}
 & & & \textbf{mean} & \textbf{std} & \textbf{mean} & \textbf{std} & \textbf{mean} & \textbf{std} & \textbf{mean} & \textbf{std} & \textbf{mean} & \textbf{std} & \textbf{mean} & \textbf{std} & \textbf{mean} & \textbf{std} & \textbf{mean} & \textbf{std} \\
\hline
\multirow{12}{*}{\rotatebox[origin=c]{90}{Newman1}} & \multirow{3}{*}{2} & 0.60 & 7.03 & 1.12 & \cellcolor{gray!25}\textbf{0.74} & 0.05 & 12.88 & 0.00 & \textbf{1.14} & 0.01 & 12.88 & 0.01 & \textbf{1.15} & 0.02 & 12.88 & 0.00 & \textbf{1.14} & 0.01 \\
 &  & 0.90 & 7.61 & 1.10 & \cellcolor{gray!25}\textbf{1.41} & 0.05 & 13.21 & 0.00 & \textbf{1.80} & 0.01 & 13.21 & 0.01 & \textbf{1.81} & 0.02 & 13.21 & 0.00 & \textbf{1.81} & 0.01 \\
 &  & 0.99 & 8.19 & 1.08 & \cellcolor{gray!25}\textbf{2.08} & 0.05 & 13.55 & 0.00 & \textbf{2.47} & 0.01 & 13.55 & 0.00 & \textbf{2.48} & 0.02 & 13.55 & 0.00 & \textbf{2.48} & 0.01 \\
 \cline{2-19}
 & \multirow{3}{*}{5} & 0.60 & 11.66 & 1.71 & \cellcolor{gray!25}\textbf{0.84} & 0.04 & 15.37 & 0.01 & \textbf{1.26} & 0.01 & 19.99 & 0.00 & \textbf{1.56} & 0.30 & 15.37 & 0.01 & \textbf{1.25} & 0.01 \\
\hline
\end{tabular}
}
\end{table*}

\begin{table*}[!t]
\centering
\caption{Statistical comparison of total offline error.}
\label{tab:offline_error_stats_all}
\resizebox{\textwidth}{!}{%
\begin{tabular}{lllrrrrrrrr}
\hline
\textbf{Inst.} & \textbf{$\nu$} & \textbf{$\alpha$} &
\textbf{MOEA/D-RE\textsuperscript{(1)}} &
\textbf{MOEA/D-DIV\textsuperscript{(2)}} &
\textbf{NSGA-II-RE\textsuperscript{(3)} }&
\textbf{NSGA-II-DIV\textsuperscript{(4)} }&
\textbf{SPEA2-RE\textsuperscript{(5)}} &
\textbf{SPEA2-DIV\textsuperscript{(6)}} &
\textbf{SMS-EMOA-RE\textsuperscript{(7)}} &
\textbf{SMS-EMOA-DIV\textsuperscript{(8)}}\\
\hline
\multirow{12}{*}{\rotatebox[origin=c]{90}{Newman1}} & \multirow{3}{*}{2} & 0.60 & {\footnotesize 2$^{-}$3$^{+}$4$^{-}$5$^{+}$6$^{*}$7$^{*}$8$^{-}$} & {\footnotesize 1$^{+}$3$^{+}$4$^{*}$5$^{+}$6$^{+}$7$^{+}$8$^{+}$} & {\footnotesize 1$^{-}$2$^{-}$4$^{-}$5$^{*}$6$^{-}$7$^{*}$8$^{-}$} & {\footnotesize 1$^{+}$2$^{*}$3$^{+}$5$^{+}$6$^{*}$7$^{+}$8$^{*}$} & {\footnotesize 1$^{-}$2$^{-}$3$^{*}$4$^{-}$6$^{-}$7$^{*}$8$^{-}$} & {\footnotesize 1$^{*}$2$^{-}$3$^{+}$4$^{*}$5$^{+}$7$^{+}$8$^{*}$} & {\footnotesize 1$^{*}$2$^{-}$3$^{*}$4$^{-}$5$^{*}$6$^{-}$8$^{-}$} & {\footnotesize 1$^{+}$2$^{-}$3$^{+}$4$^{*}$5$^{+}$6$^{*}$7$^{+}$} \\
 &  & 0.90 & {\footnotesize 2$^{-}$3$^{+}$4$^{-}$5$^{+}$6$^{*}$7$^{*}$8$^{-}$} & {\footnotesize 1$^{+}$3$^{+}$4$^{*}$5$^{+}$6$^{+}$7$^{+}$8$^{+}$} & {\footnotesize 1$^{-}$2$^{-}$4$^{-}$5$^{*}$6$^{-}$7$^{*}$8$^{-}$} & {\footnotesize 1$^{+}$2$^{*}$3$^{+}$5$^{+}$6$^{*}$7$^{+}$8$^{*}$} & {\footnotesize 1$^{-}$2$^{-}$3$^{*}$4$^{-}$6$^{-}$7$^{*}$8$^{-}$} & {\footnotesize 1$^{*}$2$^{-}$3$^{+}$4$^{*}$5$^{+}$7$^{+}$8$^{*}$} & {\footnotesize 1$^{*}$2$^{-}$3$^{*}$4$^{-}$5$^{*}$6$^{-}$8$^{-}$} & {\footnotesize 1$^{+}$2$^{-}$3$^{+}$4$^{*}$5$^{+}$6$^{*}$7$^{+}$} \\
\hline
\end{tabular}
}
\end{table*}

Additionally, Table~\ref{tab:offline_error_stats_all} reports the statistical comparison of total offline error for all eight algorithms on the Newman1, P4HD, and Marvin instances under different confidence levels and dynamic change frequencies. Pairwise statistical tests confirm that the diversity-based variants (–DIV) significantly outperform their corresponding re-evaluation-based baselines (–RE). Similar performance trends are observed for the remaining mining instances.

Our results indicate that the diversity-based method significantly
outperforms the corresponding re-evaluation–based baselines across
different frequencies of dynamic changes and confidence levels in mining
instances with up to $53{,}271$ blocks. Moreover, the proposed bi-objective
formulation produces a set of solutions for any
confidence level within the considered range in a single run. As a result, solutions for new
confidence levels can be obtained directly from the same non-dominated set
without requiring additional optimization runs. Overall, these findings
demonstrate that the proposed bi-objective formulation and the dynamic
diversity-based change response mechanism are well-suited for large-scale
open-pit mine scheduling in stochastic and dynamic environments.

\section{Conclusion}
\label{sec:conclusion} 

In this paper, we investigated the dynamic chance-constrained approach to solve the open-pit mine scheduling problem with stochastic block economic values and dynamically changing resource capacities. 
We formulated the problem as a bi-objective optimization model that simultaneously maximizes the expected net present value and minimizes its standard deviation, providing a risk-aware framework for long-term mine planning. Unlike traditional chance-constrained approaches that require a fixed confidence level, the proposed formulation generates a Pareto front that captures a range of confidence levels in a single run. To address time-varying resource capacities, we employed a diversity-based change response mechanism that combines solution repair with the introduction of randomly generated feasible solutions. We evaluated the proposed method against re-evaluation–based baseline change response mechanisms using four multi-objective evolutionary algorithms. Experimental results demonstrate that the proposed dynamic variants consistently outperform the baselines across all mining instances, confidence levels, and dynamic change frequencies, with MOEA/D-DIV and SMS-EMOA-DIV achieving the strongest performance. Overall, these results show that the proposed approach effectively handles stochastic and dynamic variations by producing high-quality, risk-aware mine schedules. 

\section{Acknowledgement}
This work has been supported by the CNRS-Adelaide Mobility Scheme Award through grant 6021521.

\bibliographystyle{unsrtnat}
\bibliography{library}

\end{document}